\documentclass[journal]{IEEEtran}

\ifCLASSINFOpdf
\else
   \usepackage[dvips]{graphicx}
\fi
\usepackage{url}

\usepackage{graphicx}
\usepackage{color}

\usepackage{times}
\usepackage{epsfig}
\usepackage{graphicx}
\usepackage{amsmath}
\usepackage{amssymb}
\usepackage{stmaryrd}
\usepackage{comment}

\begin{document}

\title{Neural Networks, Hypersurfaces, and Radon Transforms}

\author{
 \IEEEauthorblockN{Soheil~Kolouri\IEEEauthorrefmark{1}\thanks{\IEEEauthorrefmark{1} skolouri@hrl.com}, Xuwang Yin\IEEEauthorrefmark{2}\thanks{\IEEEauthorrefmark{2} xy4cm@virginia.edu}, and Gustavo K. Rohde\IEEEauthorrefmark{2}\thanks{\IEEEauthorrefmark{2}  gustavo@virginia.edu}}\\~\\
 \IEEEauthorblockA{\IEEEauthorrefmark{1} HRL Laboratories, LLC, Malibu, CA, 90265.}\\
    \IEEEauthorblockA{\IEEEauthorrefmark{2} Imaging and Data Science Laboratory, University of Virginia, Charlottesville, VA 22908.} \thanks{This work was supported in part by NIH award GM130825.}}


\maketitle

\begin{abstract}
Connections between integration along hypersufaces, Radon transforms, and neural networks are exploited to highlight an integral geometric mathematical interpretation of neural networks. By analyzing the properties of neural networks as operators on probability distributions for observed data, we show that the distribution of outputs for any node in a neural network can be interpreted as a nonlinear projection along hypersurfaces defined by level surfaces over the input data space. We utilize these descriptions to provide new interpretation for phenomena such as nonlinearity, pooling, activation functions, and adversarial examples in neural network-based learning problems.
\end{abstract}


\IEEEpeerreviewmaketitle

\section*{Introduction}


Artificial Neural Networks (NN) have long been used as a mathematical modeling method and have recently found numerous applications in science and technology including computer vision, signal processing and machine learning \cite{lecun2015deep} to name a few. Although NN-based methods are recognized as powerful techniques, much remains to be explored about neural networks as a mathematical operator (one notable exception is the function approximation results in~\cite{baldi1989neural, cybenko1989approximation}). As a consequence, numerous doubts often accompany NN practitioners such as: how does depth add nonlinearity in a NN? What is the effect of different activation functions? What are the effects of pooling?, and many others.


This didactic note is meant to highlight an alternative interpretation of NN-based techniques and their use in supervised learning problems. By investigating the connections of machine learning classification methods with projections along hyperplanes and hypersurfaces, we highlight the links between different NN architectures and the integration geometry of linear and nonlinear Radon transforms. We then use these concepts to highlight different properties of neural networks, which may help shed light on the questions highlighted above, as well as potentially provide a path for novel studies and developments. For brevity and to reduce pre-requisites, the derivations presented fall short of rigorous mathematical proofs. The Python code to reproduce all of the figures used here is available at \url{https://github.com/rohdelab/radon-neural-network}.



\section*{Statistical regression and classification}
 Let $X$ be a compact domain of a manifold in Euclidean space (the space corresponding to input digital data) and let $h: X \rightarrow Y$, with $Y \in \mathbb{R}^{K}$ represent a map (oracle) which ascertains outputs (e.g. labels) to input data (random variable) $x \in X$. In learning problems, $y \in \mathbb{R}^{K}$ is usually a vector for which the value of the $k^{\mbox{th}}$ element represents the probability that the sample $x$ belongs to the $k^{\mbox{th}}$ class, although other regression problems can also be formulated with the same approach. 

\begin{figure*}[t]
    \centering
    \includegraphics[width=\linewidth]{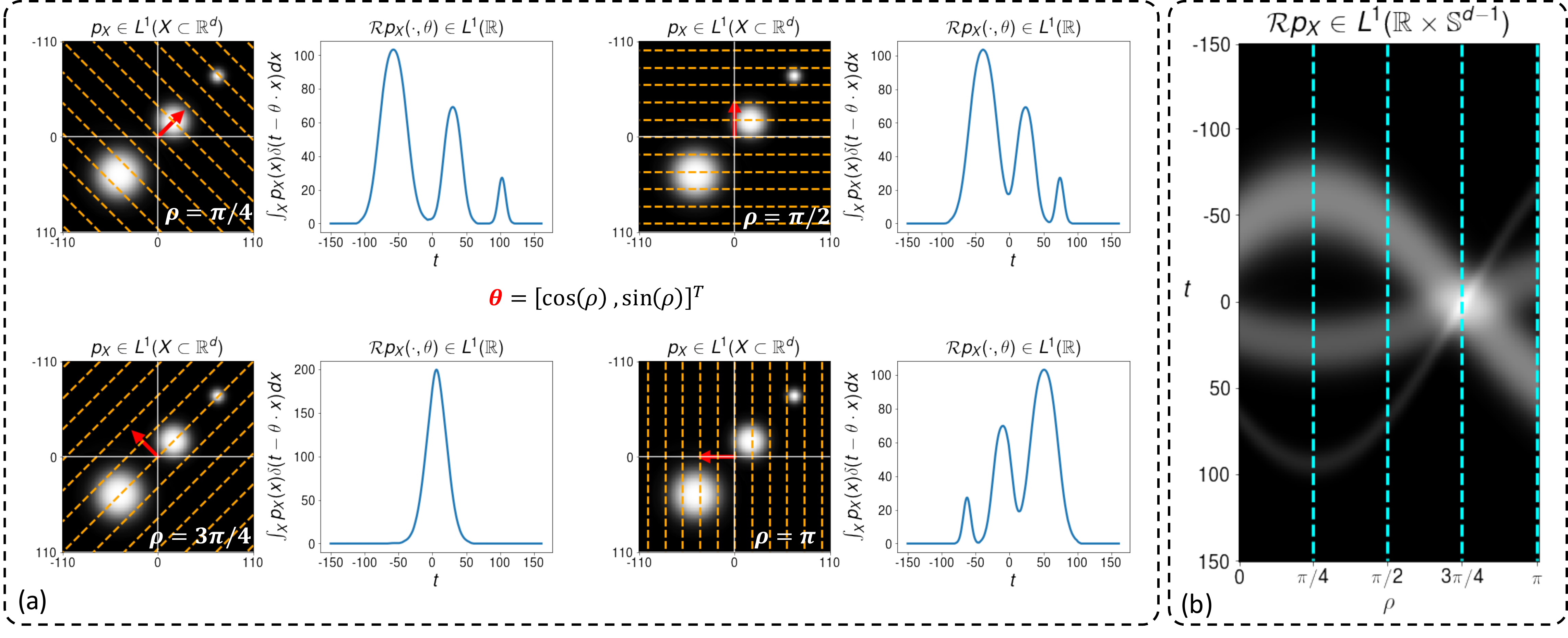}
    \caption{A visualization of the Radon transform and distribution slices. Panel (a) shows the distribution $I$, $\theta$ as a red arrow, the integration hyperplanes $H(t,\theta)$ (shown as orange lines for $d=2$), and the slices/projections $\mathcal{R}I(\cdot,\theta)$ for four different $\theta$s. Panel (b) shows the full sinogram $\mathcal{R}I$ (i.e., Radon transform), where the dotted blue lines indicate the slices shown in Panel (a). }
    \label{fig:radon_slices}
\end{figure*}

Omitting here a measure theoretic formulation (see \cite{cucker2002mathematical} for a more complete development) let $p_X, p_Y$, and $p_{X,Y} \in L_1$ (space of absolutely integrable functions) define the probability density functions (PDFs) for random variables $X$, $Y$, and $(X,Y)$,  respectively. Now utilizing a technique often used in the theoretical physics community \cite{gillespie1983theorem}, known as random variable transformation (RVT), we can write the PDF of the output $p_Y$ as a function of $p_X$ via:
\begin{equation}
	p_Y(y) = \int_{X}p_X(x) \delta(y-h(x))dx,
    \label{eq:RVT}
\end{equation}
where $\delta$ is the standard Dirac distribution. See the supplementary material for a derivation. The same transformation of random variables technique can be used to derive
\begin{equation}
	p_{f_\theta}(z) = \int_{X}p_X(x) \delta(z-f_\theta(x))dx
    \label{eq:p_F}
\end{equation}
and 
\begin{equation}
	p_{f_\theta,Y}(z,y) = \int_{X}p_X(x) \delta(y-h(x))\delta(z-f_\theta(x))dx. 
\end{equation}

The goal in a regression task is to estimate $f_\theta$ so that it accurately `predicts' the dependent variable $y$ for each input $x$. In other words, we wish to find $f_{\theta} \sim h$ over the distribution of the input space. To that end ``goodness of fit" measures are used to fit a model $f_{\theta}$ to given labeled data (supervised learning). One popular model is to find $\theta$ that minimizes the discrepancy between $y_n$ and $f_\theta(x_n)$ according to a dissimilarity measure $\mathcal{L}$:
\begin{equation}
	\min_\theta \sum_{n=1}^{N} \mathcal{L}(y_n,f_\theta(x_n)),
\end{equation}
which can be interpreted in relation to random variables $Y = h(X)$ and $f_\theta$ and their respective distributions. For instance, the cross entropy minimization strategy $\frac{-1}{N}\sum_{k=1}^{N} y_k\cdot\log(f_\theta(x_k))$ can be viewed as an estimate of $\mathbb{E}_{x\sim p_X}\left( h(x)\cdot\log(f_\theta(x))\right)$, which is equivalent to minimizing the KL-divergence between $p_Y$ and $p_{f_\theta}$. 

Next, we consider the formulations for the standard Radon transform, and its generalized version and demonstrate a connection between this transformation and the statistical learning concepts reviewed above.

\subsection{ Radon transform}  
The standard Radon transform, $\mathcal{R}$, maps distribution $p_X$ 
to the infinite set of its integrals over the hyperplanes of $\mathbb{R}^d$ and is defined as, 
\begin{eqnarray}
\mathcal{R} p_X(t,\theta):=\int_{X} p_X(x)\delta(t-x\cdot\theta)dx,
\label{eq:radon}
\end{eqnarray}
where $\delta$ is the one-dimensional Dirac delta function.
For $\forall\theta\in \mathbb{S}^{d-1}$ where $\mathbb{S}^{d-1}$ is the unit sphere in $\mathbb{R}^{d}$, and $\forall t \in \mathbb{R}$. 
Each hyperplane can be written as:
\begin{equation}
    H(t,\theta)=\{x\in \mathbb{R}^d |x\cdot\theta=t\}
    \label{eq:hyperplanes}
\end{equation}
which alternatively could be thought as the level set of the function 
$g(x,\theta)=x\cdot\theta=t$. For a fixed $\theta$, the integrals over all hyperplanes orthogonal to $\theta$ define a continuous function, $\mathcal{R}p_X(\cdot,\theta):\mathbb{R}\rightarrow\mathbb{R}$, that is a projection/slice of $p_X$. We note that the Radon transform is more broadly defined as a linear operator $\mathcal{R}: L_1(\mathbb{R}^d)\rightarrow L_1(\mathbb{R}\times \mathbb{S}^{d-1})$, where $L_1(X):=\{ I:X \rightarrow \mathbb{R} | \int_{X} |I(x)|dx \leq \infty\}$.
Figure \ref{fig:radon_slices} provides a visual representation of the Radon transform, the integration hyper-planes $H(t,\theta)$ (i.e., lines for $d=2$), and the slices $\mathcal{R}p_X(\cdot,\theta)$.
    
The Radon transform is an invertible linear transformation (i.e. linear bijection). The inverse of the Radon transform denoted by $\mathcal{R}^{-1}$ is defined as:
\begin{eqnarray}
p_X(x)&=&\mathcal{R}^{-1}(\mathcal{R}p_X(t,\theta))\nonumber\\&=& \int_{\mathbb{S}^{d-1}} (\mathcal{R}p_X(\cdot,\theta)*\eta(\cdot))\circ (x\cdot\theta)d\theta
\end{eqnarray}
where $\eta(.)$ is a one-dimensional high-pass filter with corresponding Fourier transform $\mathcal{F}\eta(\omega)\approx c|\omega|^{d-1}$ (it appears due to the Fourier slice theorem, see the supplementary material) and `$*$' denotes the convolution operation. The above definition of the inverse Radon transform is also known as the filtered back-projection method, which is extensively used in image reconstruction in the biomedical imaging community. Intuitively each one-dimensional projection/slice, $\mathcal{R}p_X(\cdot,\theta)$, is first filtered via a high-pass filter and then smeared back into $X$ along $H(\cdot,\theta)$ to approximate $p_X$. The filtered summation of all smeared approximations then reconstructs $p_X$. Note that in practice acquiring infinite number of projections is not feasible therefore the integration in the filtered back-projection formulation is replaced with a finite summation over projections.

\subsubsection{Radon transform of empirical PDFs} 
In most machine learning applications one does not have direct access to the actual distribution of the data but to its samples, $x_n\sim p_X$. In such scenarios the empirical distribution of the data is used as an approximation for $p_X$:
\begin{eqnarray}
p_X(x)\approx \hat{p}_X(x)= \frac{1}{N}\sum_{n=1}^N \delta(x-x_n)
\label{eq:edf}
\end{eqnarray} 
where $\delta$ is the Dirac delta function in $\mathbb{R}^d$. 
Then, it is straightforward to show that the Radon transform of $\hat{p}_X$ is:
\begin{eqnarray}
\mathcal{R}\hat{p}_X(t,\theta)= \frac{1}{N}\sum_{n=1}^N \delta(t-x_n\cdot\theta)
\label{eq:rtEmp}
\end{eqnarray} 


\noindent See supplementary material for detailed derivations of Equations \eqref{eq:rtEmp}.
Given the high-dimensional nature of estimating density $p_X$ in $\mathbb{R}^d$ one requires large number of samples. The projections/slices of $p_X$, $\mathcal{R}p_X(\cdot,\theta)$, however, are one dimensional and therefore it may not be critical to have large number of samples to estimate these one-dimensional densities. 

\begin{figure}
    \centering
    \includegraphics[width=\columnwidth]{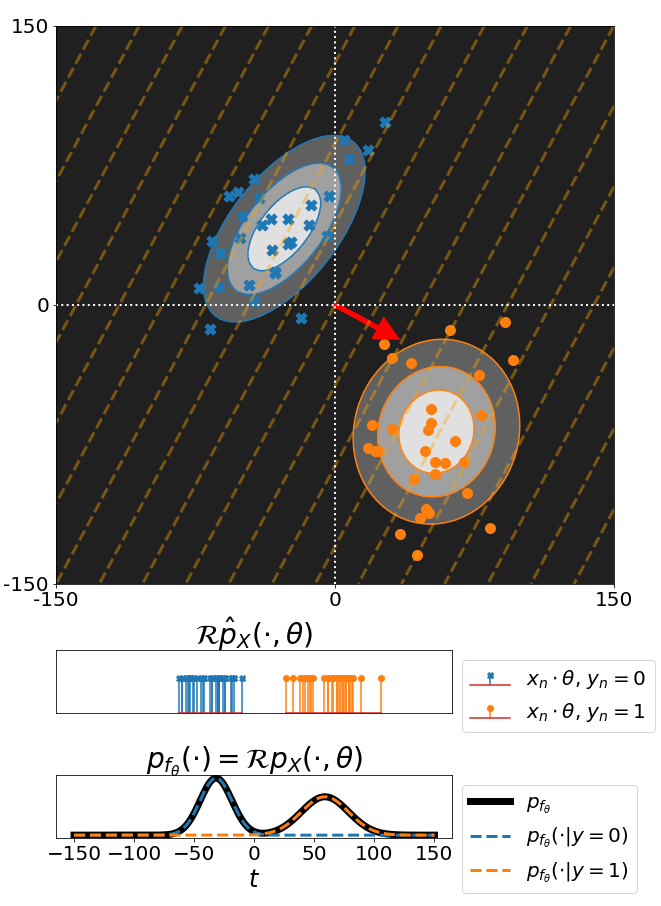}
    \caption{The linear classifier slices the distribution of the data $p_X$ at an optimal $\theta$, for which the data is best discriminated. Therefore, one can think of the distribution of the output of the classifier as a slice of the Radon transform of the distribution $p_X$.}
    \label{fig:linear_classifier}
\end{figure}

\subsubsection{Linear classification and the Radon transform}  

Now, let us consider the supervised learning of a linear binary classifier. Given the data samples $\{x_n\sim p_X\}_{n=1}^N$ and their corresponding labels $\{y_n\in\{0,1\}\}_{n=1}^N$, the task is to learn a linear function of the input samples, $f_\theta(x)=\theta\cdot x$ for $\theta\in\mathbb{S}^{d-1}$ such that,
$$
\theta\cdot x \underset{y=0}{\overset{y=1}{\gtrless}} b.
$$
Many methods exist to obtain the optimal $\theta$, e.g., Support Vector Machines or Logistic Regression. While the projection $f_\theta(x)=\theta\cdot x$ is applied to each sample, we can consider $f_\theta(\cdot)$ as an operator and inquire about the distribution $p_{f_\theta}$. Here $p_{f_\theta}(z)$ is the density of $z=f_\theta(x)$ when $x\sim p_X$. One can clearly see that $p_{f_\theta}$ corresponds to a slice of the input distribution $p_X$ with respect to $\theta\in\mathbb{S}^{d-1}$, hence there is a natural relationship between the Radon transform and linear classification. Figure \ref{fig:linear_classifier} depicts this phenomenon.

\subsection{ Generalized Radon transform} 
Generalized Radon transform (GRT) extends the original idea of the classic Radon transform introduced by J. Radon \cite{radon1917uber} from integration over hyperplanes of $\mathbb{R}^d$ to integration over hypersurfaces 
\cite{ehrenpreis2003universality,kuchment2006generalized} (i.e. $(d-1)$-dimensional manifolds). GRT  has various applications including Thermoacoustic Tomography (TAT), where the hypersurfaces are spheres, and Electrical Impedance Tomography (EIT), where integration over hyperbolic  surfaces appear.

To formally define the GRT, we introduce a function $g$ defined on $X \times (\mathbb{R}^n \backslash \{ 0 \})$ with $X \subset \mathbb{R}^d$. We say that $g$ is a \emph{defining function} when it satisfies the four conditions below:

\begin{enumerate}
    \item $g(x,\theta)$ is a real-valued $C^\infty$ function on $X \times (\mathbb{R}^n \backslash \{ 0 \})$
    \item $g(x, \theta)$ is homogeneous of degree one in $\theta$, \textit{i.e.}
    \begin{equation*}
        \forall \lambda \in \mathbb{R},\; g(x, \lambda \theta) = \lambda g(x, \theta)
    \end{equation*}
    \item $g$ is non-degenerate in the sense that $d_x g(x, \theta) \neq 0$ in $X \times \mathbb{R}^n \backslash \{ 0 \}$
    \item The mixed Hessian of $g$ is strictly positive, \textit{i.e.} 
    \begin{equation*}
        \text{det}\left( \frac{\partial^2 g}{\partial x^i \partial \theta^j} \right) > 0
    \end{equation*}
\end{enumerate}


For a given defining function, $g$, the generalized Radon transform is a linear operator  $\mathcal{G}:L^1(X)\rightarrow L^1(X \times \Omega_\theta)$, where $\Omega_\theta\subseteq (\mathbb{R}^n \backslash \{0\})$ and is defined as:
\begin{eqnarray}
\mathcal{G} p_X(t,\theta):=\int_{X} p_X(x)\delta(t-g(x,\theta))dx
\label{eq:grt}
\end{eqnarray} 
From a geometrical perspective and for a fixed $t$, $\mathcal{G}p_X(t,\theta)$ is the integral of $p_X$ along the hypersurface $H(t,\theta)=\{x\in X|g(x,\theta)=t\}$. Note that the classic Radon transform is a special case of the generalized Radon transform where $g(x,\theta)=x\cdot\theta$. 

The investigation of the sufficient and necessary conditions for showing the injectivity of GRTs is a long-standing topic \cite{ehrenpreis2003universality,kuchment2006generalized}. The conditions 1-4 for a defining function, $g$, enumerated in this section, are necessary conditions for injectivity but not sufficient. 
Though the topic related to inversion of the GRT is beyond the scope of this article, an inversion approach is given in \cite{uhlmann2003inside}.

Here, we list a few examples of known defining functions that lead to injective GRTs.  The circular defining function, $g(x,\theta) = \|x-r*\theta\|_2$ with $r\in\mathbb{R}^+$ and $\Omega_\theta = \mathbb{S}^{d-1}$ was shown to provide an injective GRT \cite{kuchment2006generalized}. More interestingly, homogeneous polynomials with an odd degree also yield an injective GRT \cite{ehrenpreis2003universality}, \textit{i.e.}
    $g(x,\theta) = \sum_{|\alpha| = m} \theta_\alpha x^\alpha$,
where we use the multi-index notation $\alpha = (\alpha_1, \dots, \alpha_{d_\alpha}) \in \mathbb{N}^{d_\alpha}$, $|\alpha| = \sum_{i=1}^{d_\alpha} \alpha_i$, and $x^\alpha = \prod_{i=1}^{d_\alpha} x_i^{\alpha_i}$. Here, the summation iterates over all possible multi-indices $\alpha$, such that $|\alpha| = m$, where $m$ denotes the degree of the polynomial and $\theta_\alpha \in \mathbb{R}$. The parameter set for homogeneous polynomials is then set to $\Omega_\theta=\mathbb{S}^{d_\alpha-1}$.  We can observe that choosing $m=1$ reduces to the linear case $ g(x,\theta)=x\cdot\theta$, since the set of the multi-indices with $|\alpha|=1$ becomes $\{ (\alpha_1, \dots, \alpha_d); \alpha_i = 1 \text{ for a single } i\in \llbracket 1, d \rrbracket, \text{ and } \alpha_j = 0, \quad \forall j \neq i\}$ and contains $d$ elements.


 \begin{figure*}[t!]
\centering
\includegraphics[width=\linewidth]{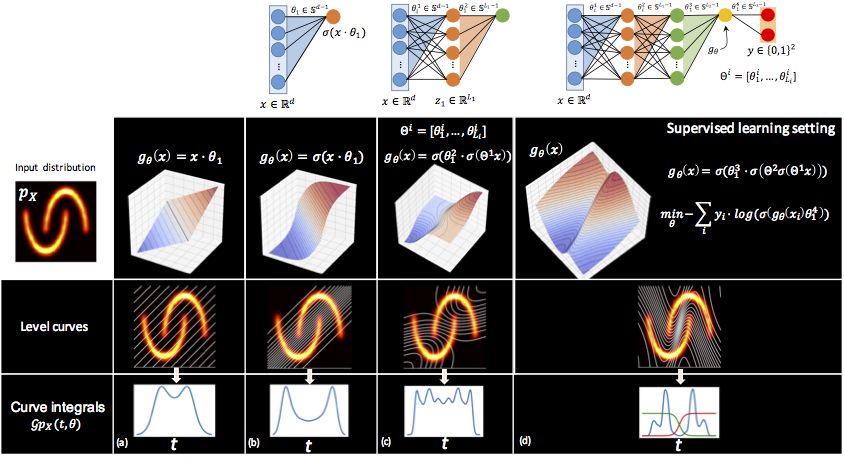}
\caption{Curve integrals for the half-moon dataset for a random linear projection, which is equivalent to a slice of linear Radon transform (a), for one layer perceptron with random initialization, which is isomorphic to the linear projection (b), for a two-layer perceptron with random initialization (c), and for a trained multi-layer perceptron (d). The weights of the all perceptrons are forced to be normalized so that $\|\theta_i^j\|=1$.}
\label{fig:perceptron}
\end{figure*}

\section*{ neural networks and the generalized Radon transform}

To illustrate the relationship between deep neural networks and the generalized Radon transform we start by describing the link between perceptrons and the standard Radon transform.

\subsection{Single perceptron}

Let $z=\sigma(\theta\cdot x)$, with $\|\theta\|=1$, define a perceptron for input data $x\sim p_X$, where we dissolved the bias, $b$, into $\theta$. Treating $z\sim p_Z$ as a random variable and using RVT, it is straightforward to show that $p_{Z}$ is isomorphic to a single slice of $p_X$, $\mathcal{R}p_X(t,\theta)$, when $\sigma$ is invertible (see supplementary material for a proof). The isomorphic relationship provides a fresh perspective on perceptrons, stating that the distribution of the perceptron's output, $f_\theta(x)$, is equivalent to integration of the original data distribution, $p_X$, along hyperplanes $H(t,\theta)=\{x|x\cdot\theta=\sigma^{-1}(t)\}$ (see Equation \eqref{eq:hyperplanes}). In addition, one can show that the distribution of the output of a perceptron is equal to the generalized Radon transform with $g(x,\theta)=f_\theta(x)$,

\begin{eqnarray}
p_{f_\theta}(z)=\mathcal{G}p_X(z,\theta)=\int_X p_X(x)\delta(z-\sigma(x\cdot\theta))dx.
\end{eqnarray}

An important and distinctive point here is that here we are interested in the distribution of the output of a perceptron,  $\mathcal{G}p_X(z,\theta)$, and its relationship to the original distribution of the data, $p_X$, as opposed to the individual responses of the perceptron, $z_n=g(x_n,\theta)$. Columns (a) and (b) in Figure \ref{fig:perceptron} demonstrate the level sets (or level curves since $d=2$) and the line integrals for $g(x,\theta)=x\cdot\theta$ and $g(x,\theta)=\sigma(x\cdot\theta)$, where $\theta\in\mathbb{S}^1$. Note that samples that lay on the same level set will be mapped to a fixed projection (a constant value $z$). In other words, the samples that lay on the same level sets of $g(x,\theta)$ are indistinguishable in the range of the perceptron. Next we discuss the case of having multiple perceptrons.


\subsection{Multilayer (Deep) neural networks}
To obtain a hierarchical (multilayer) model, the concept of a perceptron can be applied recursively. As before, let $\Theta^1$ and $\Theta^2$ correspond to two matrices whose rows contain a set of projection vectors (different $\theta$'s in the preceding section): e.g. $\Theta^{1} = [\theta_{1}^{(1)^T}, \theta_{2}^{(1)T}, \cdots]$ where $\theta_{1}^{(1)^T}$ is the transpose of projection vector corresponding to the first node/perceptron in layer 1. A two layer NN model can be written as $\sigma(\Theta^2\sigma(\Theta^1) x)$. Expanding the idea further, we then may define a general formula for a $K$-layer NN as 
\begin{equation}
g(x,\theta)= \sigma(\theta_1^K\cdot \sigma(\Theta^{K-1}\sigma(\Theta^{K-2}(...\sigma(\Theta^1x)))))
\label{eq:mlp}
\end{equation}
Note that $\theta_1^K$ above refers to a column vector which collapses the output of the neural network to one node and that $\Theta^k=[\theta_1^k,...,\theta_{L_k}^k]$ where $L_k$ is the number of neurons in the $k$'th layer of a deep neural network. 

Now, let $\sigma$ be a Lipschitz continuous nonlinear activation function. Its self-composition is therefore also Lipschitz continuous. For invertible activation functions $\sigma$, and for $\Theta^{k}$ square and invertible, the gradient of a multi-layer perceptron in equation \eqref{eq:mlp}  does not vanish in any compact subset of $\mathbb{R}^d$ and therefore the level sets are well-behaved.  Therefore, from the definition in \eqref{eq:RVT} we have that the distribution over the output node $p_Y(y)$ could be considered as a slice of the generalized Radon transform of $p_X$ evaluated at $\theta$: $p_Y(y) = \mathcal{G}p_X(t,\theta)$ with 
\begin{equation}
    \mathcal{G}p_X(t,\theta) = \int_{X}p_X(x)\delta(t-\sigma(\theta_1^K\cdot \sigma(\Theta^{K-1}...\sigma(\Theta^1x)))) dx \notag
\end{equation}
Figure \ref{fig:perceptron} columns (c) and (d) demonstrate the level sets and the line integrals of $p_X(x)$ using a multi-layer perceptron as $g_\theta$. Column (c) is initialized randomly and column (d) shows $g_\theta$ after the network parameters are trained in a supervised classification setting to discriminate the modes of the half-moon distribution. It can be seen that after training, the level sets, $H(t,\theta)$, only traverse a single mode of the half-moon distribution, which indicates that the samples from different modes are not projected onto the same point (i.e. the distribution is not integrated across different modes). It also readily becomes apparent the facility with which neural networks have to generate highly nonlinear functions, even with relatively few parameters (below we compare these to other polynomials). We note that with just one layer, NN's can form nonlinear decision boundaries, as the superposition of surfaces formed by $\sigma(\theta_1^1 \cdot x)+\sigma(\theta_2^1 \cdot x) +\cdots$ can add curvature to the resulting surface. Note also that generally speaking, the integration streamlines (hypersurfaces for higher dimensional data) have the ability to become more curved (nonlinear) as the number of layers increases. 
 



\begin{figure}[t!]
\centering
\includegraphics[width=0.9\columnwidth]{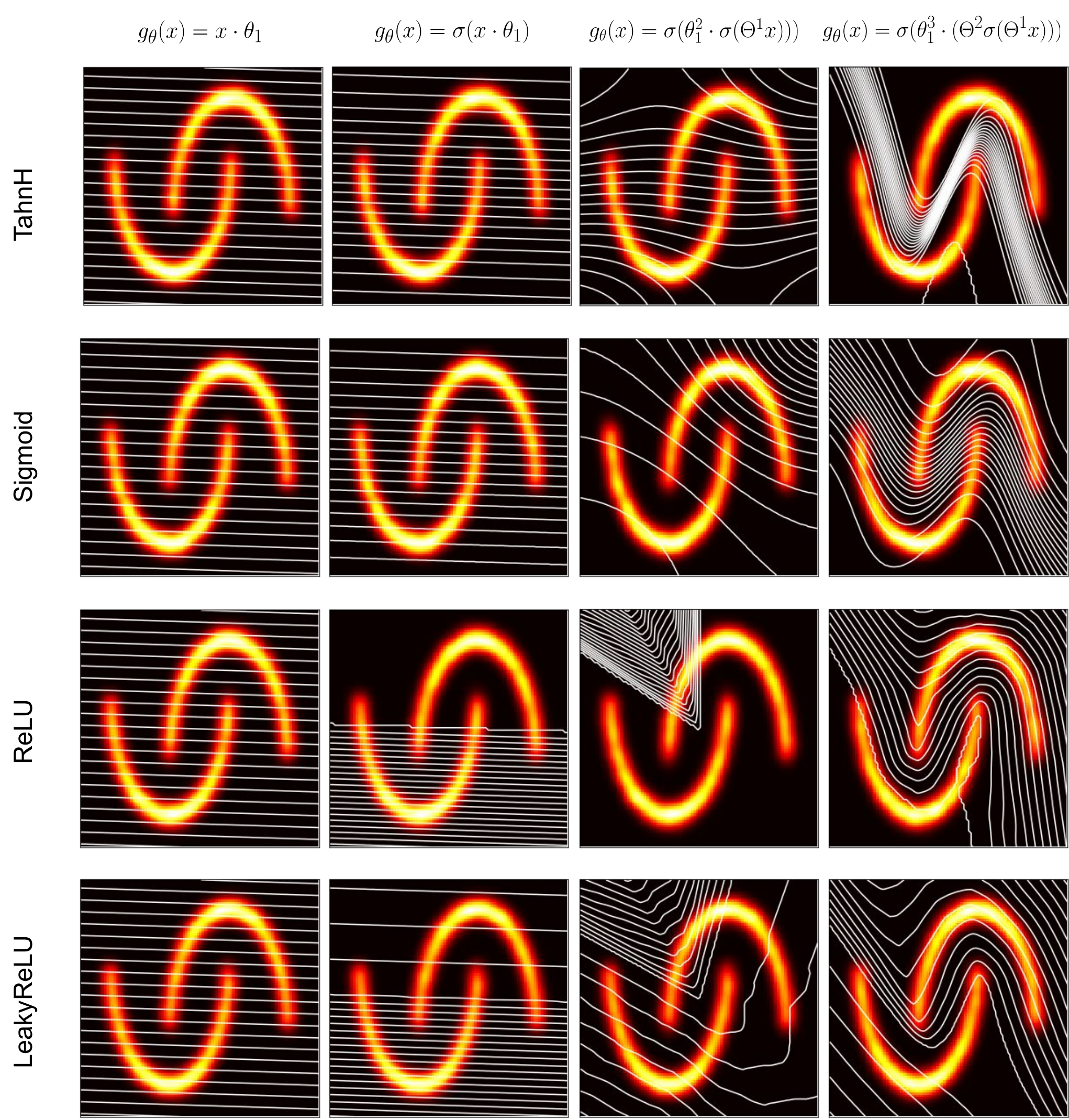}
\caption{Level curves of nodes introduced by different activation functions. Parameters $\theta_1 \in \mathbb{R}^2, \Theta^1 \in \mathbb{R}^{50\times 2}, \theta_1^2\in \mathbb{R}^{50}$ are randomly initialized (with the same seed) for the first three column demonstrations. Parameters for the last column  $\Theta^1 \in \mathbb{R}^{50\times 2}, \Theta^2 \in \mathbb{R}^{100\times 50}, \theta^3_1 \in \mathbb{R}^{100}$ are optimized by minimizing a misclassification loss.}
\label{fig:activations}
\end{figure}
\subsection*{Activation functions}


It has been noted recently that NN's (e.g. convolutional neural networks) can at times work better when $\sigma$ is chosen to be the rectified linear unit (ReLU) function, as opposed the sigmoid option~\cite{nair2010rectified, glorot2011deep, krizhevsky2012imagenet}. The experience has encouraged others to try different activation functions such as the `leaky'-ReLU~\cite{maas2013rectifier}. While theory describing which type of activation function should be used with which type of learning problem is yet to emerge, the interpretation of NN's as nonlinear projections can help highlight the differences between activation function types. Specifically, Figure~\ref{fig:activations} can help visualize the effects of different activation functions on the integration geometry over the input data space $X$. 

First note that the ReLU is a non-invertible map, given that negative values all map to zero. This will cause the surface generated by a perceptron constructed with ReLU to have a region over $X$ which is flat, whereby all points in that region are integrated and mapped to the same value (zero) in the output space. This ability may provide ReLU neural networks with the flexibility to construct adaptable characteristic function-type models for different regions in the data space. Although, the outcome of the optimization procedure will dictate whether such regions would emerge in the final model. Finally, note that both ReLU and the leaky-ReLU activation functions  contain non-differentiable points, which are also imparted on the surface function (hence the sharp `kinks' that appear over iso-surfaces lines).

 \begin{figure*}[t!]
\centering
\includegraphics[width=\linewidth]{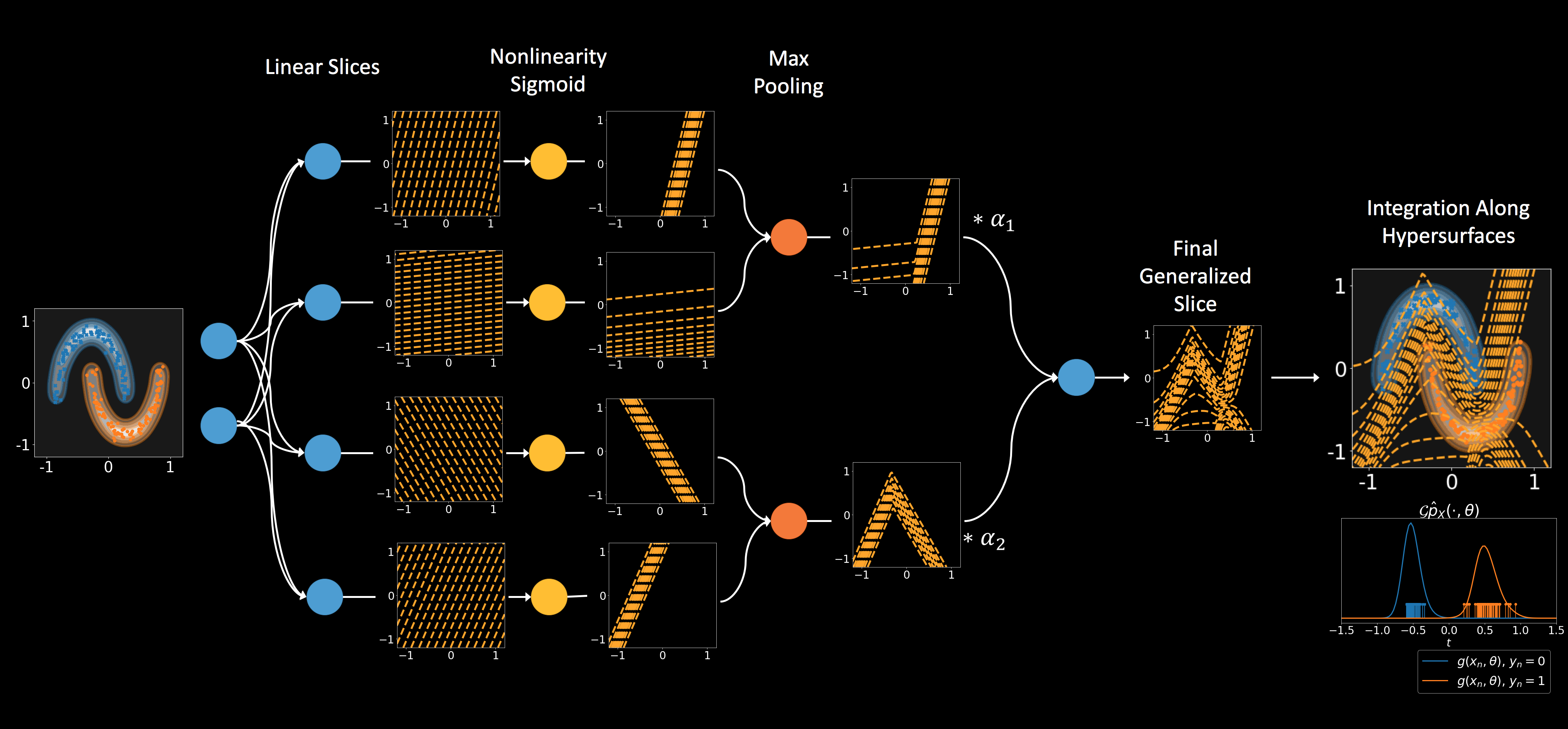}
\caption{Demonstration of max pooling operation. The level surfaces corresponding to perceptron outputs for a given input sample $x$ are selected for maximum response (see text for more details).}
\label{fig:max_pooling}
\end{figure*}

\subsection*{Pooling}

Pooling (e.g. average or maximum sampling) operations are typically used in large neural networks, especially the CNN kind. The reasons are often practical, as subsampling can be used as a way to control and reduce the dimension (number of parameters) of the model. Another often stated reason is that pooling can also add a certain amount of invariance (e.g. translation) to the NN model. In the case of average pooling, it is clear that the operation can also be written as a linear operator $\Theta^k$ in equation \eqref{eq:mlp} where the pooling operation can be performed by replacing a particular row of $\Theta^k$ by the desired linear combination between two rows of $\Theta^k$, for example. `Max'-pooling on the other hand selects the maximum surface value (perceptron response), over all surfaces (each generated by different perceptrons) in a given layer.  Figure \ref{fig:max_pooling} shows a graphical description of the concept, though it should be noted that as defined above, the integration lines are not being added, rather the underlying `level' surfaces. 

 \subsection*{Adversarial examples}
 
\begin{figure}[t!]
\centering
\includegraphics[width=0.8\columnwidth]{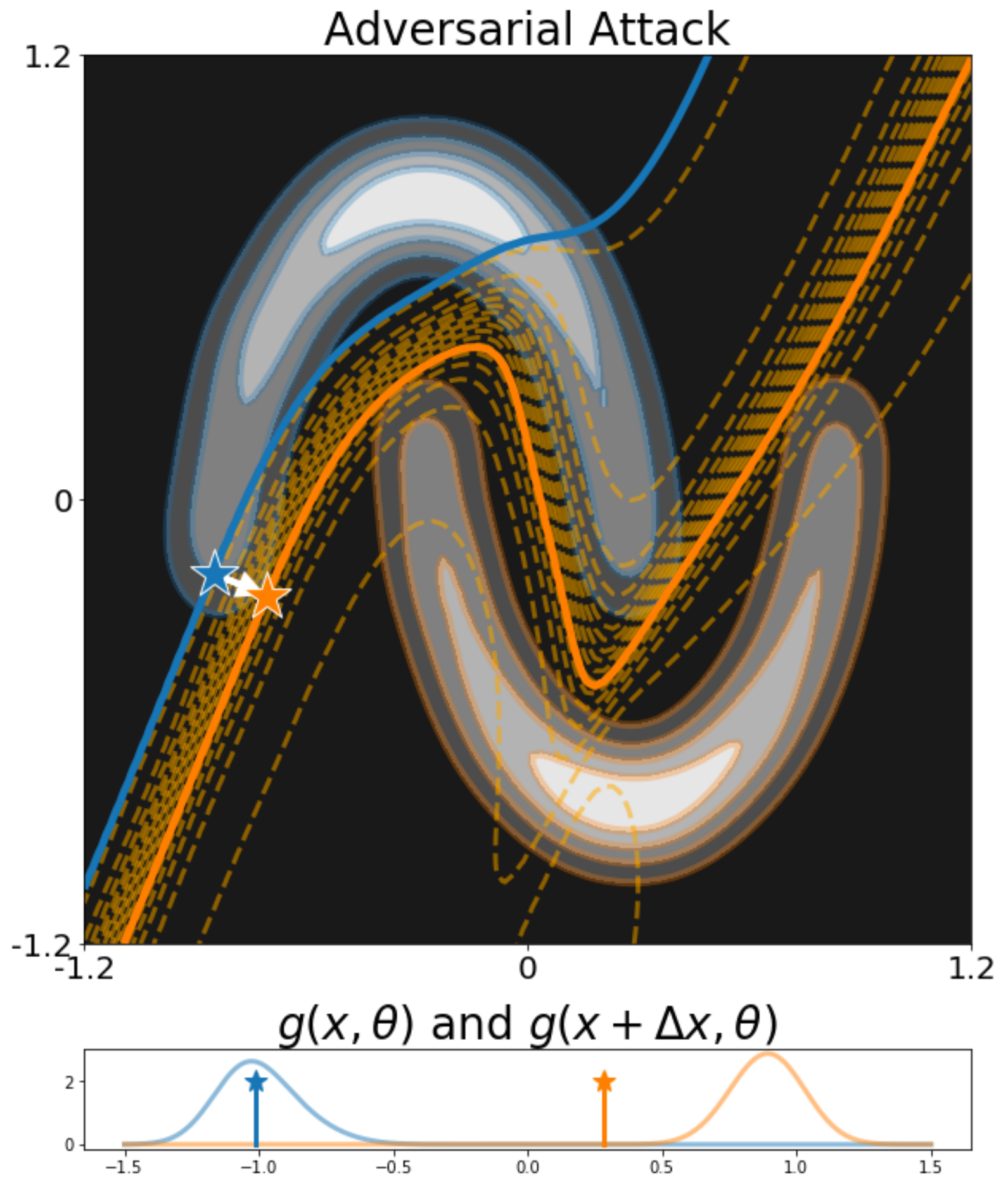}
\caption{Adversarial perturbations lead to a shift between hypersurfaces.}
\vspace{-.2in}
\label{fig:adversarial}
\end{figure}

 It has often been noted that highly flexible nonlinear learning systems such as CNN's can be `brittle' in the sense that a seemingly small perturbation of the input data can have cause the learning system to produce confident, but erroneous, outputs. Such perturbed examples are often termed as adversarial examples. Figure \ref{fig:adversarial} utilizes the integral geometric perspective described above to provide a visualization of how neural networks (as well as other classification systems) can be fooled by small perturbations. To find the minimum displacement that could cause misclassificaiton, using the blue point as the starting point $x_0$, we perform gradient ascent $ {x}_{n+1}= {x}_{n}+\gamma \nabla g(x_n, \theta)$, until we reach the other side of the decision boundary (which is indicated by the orange point).  We limit the magnitude of the displacement small enough so that the two points belong to the same distribution. However, once integrated along the isosurfaces corresponding to the NN drawn in the figure, due to the uneven curvature of the corresponding surface, the two points are projected onto opposite ends of the output node, thus fooling the classifier to make a confident, but erroneous, prediction.

\begin{figure}[t!]
\centering
\includegraphics[width=0.9\columnwidth]{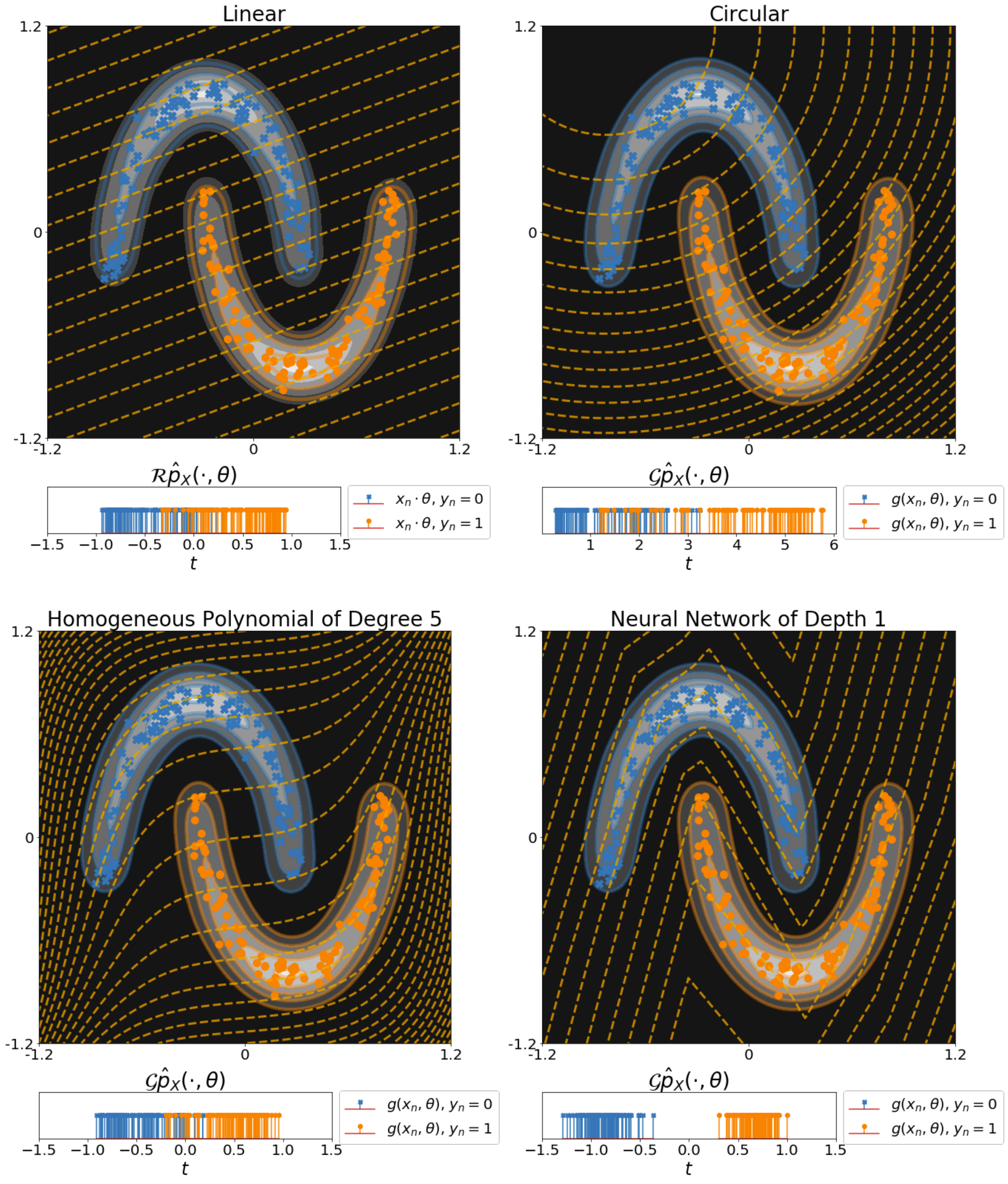}
\caption{The level curves (i.e. hyperplanes and hypersurfaces), $H(\cdot,\theta)$, with optimally discriminant $\theta$, for different defining functions, namely linear (i.e., the standard Radon transform), circular, homogeneous polynomial of degree 5, and a multi-layer perceptron with leaky-ReLU activations.}
\vspace{-.2in}
\label{fig:spherical}
\end{figure}




\section*{Summary and future directions}

In this note we explored links between Radon transforms and Artificial Neural Networks to study the properties of the later as an operator on the probability density function $p_X$ associated with a given learning problem. More specifically, it can be shown that the probability density function associated with any output node of a neural network (or a single perceptron within a hierarchical NN) can be interpreted as a particular hyperplane integration operation over the distribution $p_X$. This interpretation has natural connections with the $N$-dimensional generalized Radon transforms, which similarly proposes that a high-dimensional PDF can be represented by integrals defined over linear or nonlinear hyperplanes in $X$. The projections can be linear (in the case of simple linear logistic regression) or nonlinear in which case the projection are computed over nonlinear hyperplanes. The analogy has limitations, however, given that depending on the number of nodes in each layer, as well as the choice of activation function, the conditions for the generalized Radon transforms in \cite{uhlmann2003inside} (i.e. invertibility, homogeneity, etc) may not be satisfied with specific neural network architectures.

Despite these limitations, the analogy is useful to provide a mechanistic understanding of NN operators, and it may also be useful as a path to study the effect of different neural network architectures and related concepts (number of layers, number of nodes in each layer, choice of activation function, recurrency, etc.) as well as to provide ideas for alternative models. For example, other types of projections may be considered within a learning problem. Figure \ref{fig:spherical} compares linear projections, circular projections, a homogeneous polynomial of degree 5, and an ANN of depth 1, all trained to minimize the logistic regression cost function. While it is clear that linear and circular projections don't have enough `flexibility' to solve the separation problem, a polynomial degree of degree 5 seems to emulate the behavior of an ANN of depth 1. It is possible that in the future, the point of view provided by analyzing the nonlinear projections associated with different NN's can provide inspiration for alternative models.


\bibliographystyle{IEEEtran}
\bibliography{IEEEabrv,NN_Generalized_Radon}

\clearpage
\section{Supplementary material}

\subsection{Inverse of Radon transform}

To define the inverse of the Radon transform we start by the Fourier slice theorem. Let $\mathcal{F}_d$ be the d-dimensional Fourier transform, then the one dimensional Fourier transform of a projection/slice is:
\begin{eqnarray*}
\mathcal{F}_1(\mathcal{R}I(\cdot,\theta))(\omega)&=&\int_\mathbb{R}\mathcal{R}I(t,\theta)e^{-i\omega t}dt\\
&=&\int_\mathbb{R}\int_{\mathbb{R}^d} I(x)e^{-i\omega t}\delta(t-x\cdot \theta)dxdt\\
&=&\int_{\mathbb{R}^d} I(x)e^{-i(\omega\theta)\cdot x}dx=\mathcal{F}_dI(\omega\theta)
\end{eqnarray*}
which indicates that the one-dimensional Fourier transform of each projection/slice is equal to a slice of the d-dimensional Fourier transform in a spherical coordinate. Taking the inverse d-dimensional Fourier transform of $\mathcal{F}_dI(\omega\theta)$ in the Cartesian coordinate, $u\in\mathbb{R}^d$, would lead to the reconstruction of $I$. 

\begin{eqnarray*}
I(x)&=&\mathcal{F}_d^{-1}(\mathcal{F}_dI(u))\\
&=& \int_\mathbb{R}\int_{\mathbb{S}^{d-1}} \mathcal{F}I(\omega\theta)e^{i\omega\theta\cdot x}|\omega|^{d-1} c(\theta)d\theta dt
\end{eqnarray*}
where,
\begin{equation*}
    c(\theta)=\sin^{d-2}(\theta_1)\sin^{d-3}(\theta_2)...\sin(\theta_{d-2})
\end{equation*}
where $\theta=[\theta_1,...,\theta_d-1]$, and $c(\theta)$ is often approximated as a small angle-independent constant, $c$.

\subsection{The RVT theorem}

Here we show the derivations for Equation \eqref{eq:RVT}. Recall that $h(x)=y$ is the true map (oracle) from the data samples to their corresponding labels. Let $g:\mathbb{R}\rightarrow \mathbb{R}$ be a real function, then we can write $g(y)=g(h(x))$. By definition, the average of the quantity on the left with respect to $y$ should be equal to the average of the quantity on the right with respect to $x$, and we can write:
\begin{eqnarray}
\int_\mathbb{R} g(y)p_Y(y)dy&=& \int_X g(h(x))p_X(x)dx \nonumber \\
&=& \int_X\int_\mathbb{R} g(y)\delta(y-h(x))p_X(x)dydx \nonumber\\
\label{eq:rvt_proof}
\end{eqnarray}
Now let $g(y)=\delta(y-y')$ and for the left hand side of Equation \eqref{eq:rvt_proof} we have:
\begin{eqnarray*}
\int_\mathbb{R} \delta(y-y')p_Y(y)dy&=& p_Y(y')
\end{eqnarray*}
and for the right hand side of Equation \eqref{eq:rvt_proof} we have:
\begin{eqnarray*}
\int_X\int_\mathbb{R} &\delta(y-y')\delta(y-h(x))p_X(x)dydx =& \\ &\int_X p_X(x)\delta(y'-h(x))dx&
\end{eqnarray*}
which yields:
\begin{eqnarray*}
p_Y(y') = \int_X p_X(x)\delta(y'-h(x))dx,
\end{eqnarray*}
and concludes our proof the derivation. For a more complete analysis, see \cite{gillespie1983theorem}.

\subsection{Isomorphic relationship between a perceptron and a standard Radon slice }

For the perceptron, $f_\theta(x)=\sigma(x\cdot\theta)$, where $\|\theta\|=1$ and $\sigma:\mathbb{R}\rightarrow U=(0,1)$, the distribution of the output could be obtained from:
\begin{eqnarray*}
p_{f_\theta}(z)=\int_X p_X(x)\delta(z-\sigma(x\cdot\theta))dx
\end{eqnarray*}
on the other hand, the Radon slice of $p_X$ is obtained from 
\begin{eqnarray*}
\mathcal{R}p_X(t,\theta)=\int_X p_X(x)\delta(t-x\cdot\theta)dx
\end{eqnarray*}
We first show that having $\mathcal{R}p_X(\cdot,\theta)$ one can recover $p_{f_\theta}$. Let $z=\sigma(t)$, where $t\sim \mathcal{R}p_X(\cdot,\theta)$ therefore using RVT the distribution of $z$ is equal to:
\begin{eqnarray*}
p_Z(z)&=&\int_\mathbb{R} \mathcal{R}p_X(t,\theta)\delta(z-\sigma(t))dt\\
&=&\int_\mathbb{R}\int_X p_X(x)\delta(t-x\cdot\theta)\delta(z-\sigma(t))dxdt\\
&=&\int_X p_X(x)\delta(z-\sigma(x\cdot\theta))dx\\
&=& p_{f_\theta}(z)
\end{eqnarray*}
Now we show the reverse arguement. For invertible $\sigma$, let $t=\sigma^{-1}(z)$ where $z\sim p_{f_\theta}$, then we can obtain the distribution of $t$ from:
\begin{eqnarray*}
p_T(t)&=&\int_U p_{f_\theta}(z)\delta(t-\sigma^{-1}(z))dz\\
&=&\int_U\int_X p_X(x)\delta(z-\sigma(x\cdot\theta))\delta(t-\sigma^{-1}(z))dxdz\\
&=&\int_X p_X(x)\delta(t-\sigma^{-1}(\sigma(x\cdot\theta)))dx\\
&=&\int_X p_X(x)\delta(t-x\cdot\theta)dx\\
&=& \mathcal{R}p_X(t,\theta)
\end{eqnarray*}
therefore the two distributions, $\mathcal{R}p_X(\cdot,\theta)$ and $p_{f_\theta}$, are isomorphic.  
\end{document}